\documentclass[conference]{IEEEtran}
\IEEEoverridecommandlockouts
\usepackage{cite}
\usepackage{amsmath, amssymb, amsfonts}
\usepackage{algorithmic}
\usepackage{graphicx}
\usepackage{textcomp}

\usepackage{booktabs}
\usepackage{multirow}
\usepackage{array}
\usepackage{xcolor}
\usepackage{pifont}
\usepackage{caption}
\usepackage{subcaption}
\newcolumntype{P}[1]{>{\centering\arraybackslash}p{#1}}

\usepackage{footnote}
\makesavenoteenv{tabular}

\usepackage{xcolor}
\def\BibTeX{{\rm B\kern-.05em{\sc i\kern-.025em b}\kern-.08em T\kern-.1667em\lower.7ex\hbox{E}\kern-.125emX}}

\usepackage{pifont}% http://ctan.org/pkg/pifont

\usepackage[capitalize]{cleveref}
\crefname{section}{Sec.}{Secs.}
\Crefname{section}{Section}{Sections}
\Crefname{table}{Table}{Tables}
\crefname{table}{Tab.}{Tabs.}

\usepackage{tikz}
\newcommand{\copyrighttext}{%
\footnotesize \textcopyright~ 2025 IEEE. Personal use of this material is
permitted. Permission from IEEE must be obtained for all other uses, in any
current or future media, including reprinting/republishing this material for advertising
or promotional purposes, creating new collective works, for resale or redistribution
to servers or lists, or reuse of any copyrighted component of this work in other
works.}
\newcommand{\copyrightnotice}{%
\begin{tikzpicture}[remember picture, overlay]
	\node[anchor=south,yshift=10pt] at (current page.south)
	{\fbox{\parbox{\dimexpr\textwidth-\fboxsep-\fboxrule\relax}{\copyrighttext}}};
\end{tikzpicture}%
}

\begin{document}
	\title{Minimalistic Video Saliency Prediction via Efficient Decoder \& Spatio
	Temporal Action Cues

	% A Minimalist Approach to Video Saliency Prediction: Leveraging Lightweight Decoder and Spatio Temporal Action Localization
	%Video Saliency Simplified: Leveraging Lightweight Decoder and Spatio-Temporal Action Detection

	}

	\author{ \IEEEauthorblockN{ Rohit Girmaji, Siddharth Jain, Bhav Beri, Sarthak Bansal, Vineet Gandhi}
	\IEEEauthorblockA{CVIT, IIIT Hyderabad, India\\ \{rohit.girmaji, siddharth.jain, bhav.beri\}@research.iiit.ac.in, sarthak.bansal@students.iiit.ac.in, vgandhi@iiit.ac.in}
	% \\
	% \IEEEauthorblockN{ Sarthak Bansal\IEEEauthorrefmark{2},
	%                    Vineet Gandhi\IEEEauthorrefmark{2}}
	% \IEEEauthorblockA{\IEEEauthorrefmark{2}IIIT Hyderabad, India\\
	%                   sarthak.bansal@students.iiit.ac.in, vgandhi@iiit.ac.in}
	}

	\maketitle
	\copyrightnotice

	\begin{abstract}
		% This paper introduces ViNet-S, a 36MB model built upon the ViNet architecture, which follows a minimalistic U-Net-based design. Our approach incorporates a lightweight decoder over the original ViNet, resulting in over threefold reduction in model size and the number of parameters, all without compromising performance. Also, we propose a ViNet-A (148MB) model that utilizes spatio-temporal action localization (STAL) features, unlike the traditional video saliency prediction models that rely on action classification backbones. Our empirical studies reveal that a simple ensemble of ViNet-S and ViNet-A, achieved by averaging the predicted saliency maps, attains state-of-the-art performance on nine different visual-only and audio-visual saliency prediction datasets, even without using audio cues. Remarkably, the individual models and their ensemble have fewer parameters than the latest state-of-the-art transformer-based models. Benefiting from a non-autoregressive design, all our proposed models achieve real-time performance, while the smaller ViNet-S model achieves an impressive batched runtime performance of over 1000fps.

		This paper introduces ViNet-S, a 36MB model based on the ViNet architecture with
		a U-Net design, featuring a lightweight decoder that significantly reduces
		model size and parameters without compromising performance. Additionally,
		ViNet-A (148MB) incorporates spatio-temporal action localization (STAL)
		features, differing from traditional video saliency models that use action classification
		backbones. Our studies show that an ensemble of ViNet-S and ViNet-A, by averaging
		predicted saliency maps, achieves state-of-the-art performance on three
		visual-only and six audio-visual saliency datasets, outperforming transformer-based
		models in both parameter efficiency and real-time performance, with ViNet-S
		reaching over 1000fps.
	\end{abstract}


	\section{Introduction}
	Human visual attention (HVA) enables selective focus on relevant stimuli, a
	capability that computational saliency prediction (SP) aims to replicate in
	dynamic scenes. The formal approach to addressing this task involves initially
	recording human gaze using an eye-tracking hardware device and subsequently employing
	this data as the reference point for training predictive models. SP models
	have made substantial progress over the years and have shown considerable
	benefits across a wide range of applications such as intelligent robotic
	behaviour~\cite{butko2008visual}, automated cinematic editing~\cite{moorthy2020gazed},
	human-computer interaction~\cite{chang2019salgaze, ferreira2014attentional,
	mavani2017facial, schillaci2013evaluating}, and autonomous driving~\cite{lateef2021saliency}.

	% Human visual attention (HVA) is the brain's remarkable ability to selectively focus on specific elements within the visual field while ignoring others, allowing individuals to efficiently process vast amounts of visual information by prioritizing the most relevant stimuli. Extending this capability to machines and robots represents an exciting frontier in machine learning and computer vision research. Computational Saliency Prediction (SP) endeavours to replicate HVA in dynamic scenes, enabling, for instance, a robot to exhibit intelligent behaviour by focusing on salient regions, such as a painting or a door, rather than a plain wall~\cite{butko2008visual, dang2018visual}. SP models have made substantial progress over the years and have shown considerable benefits across a wide range of applications, including automated cinematic editing~\cite{moorthy2020gazed}, human-computer interaction~\cite{chang2019salgaze, ferreira2014attentional, mavani2017facial, schillaci2013evaluating}, robotic camera control~\cite{butko2008visual}, autonomous driving~\cite{lateef2021saliency}, and video compression~\cite{ZHU2018511, hadizadeh2013saliency}.

	%
	%and more recently, transformer-based networks~\cite{tmfinet,thtdnet}, have advanced the field, albeit with increased model complexity.
	%video saliency prediction models started with recurrent neural networks ~\cite{droste2020unified,dhf1k} and then employed 3D convolutional encoder-decoder network~\cite{jain2021vinet,min2019tased}.
	%

	In the deep learning era, early SP methods used two-stream approaches~\cite{kocak2021gated,8543830}
	or recurrent networks~\cite{droste2020unified,dhf1k}, which struggled with long-range
	dependencies and spatial-temporal cues. 3D convolution-based model~\cite{jain2021vinet,
	min2019tased}
	architectures then followed, which typically utilize action classification backbones
	like S3D~\cite{s3d} pre-trained on the Kinetics dataset~\cite{kay2017kinetics}.
	ViNet~\cite{jain2021vinet}, a fully convolutional encoder-decoder, uses
	hierarchical features with UNet-like~\cite{ronneberger2015unet} skip
	connections. STSANet~\cite{stsanet} employs spatio-temporal self-attention but
	is too large for practical use. Recent approaches like TMFI-Net~\cite{tmfinet}
	and THTD-Net~\cite{thtdnet} use Video Swin Transformer for saliency prediction,
	focusing on long-range temporal dependencies.

	Prior works have also explored combining audio and visual modalities for saliency
	prediction. STAViS~\cite{tsiami2020stavis} combines spatio-temporal visual and
	auditory features with linear weighting. TSFP-Net~\cite{tsfpnet} builds a temporal-spatial
	feature pyramid, fusing audio and visual features with attention mechanisms. VAM-Net~\cite{vamnet},
	VASM~\cite{mvva} employs multi-stream and multi-modal networks to predict
	saliency maps. CASP-Net~\cite{xiong2023casp} associates video frames with sound
	sources using a two-stream encoder. Recently, DiffSal~\cite{diffsal} introduced
	a diffusion-based approach for audio-visual saliency modelling; however, it suffers
	from heightened computational complexity and substantially slower inference speeds.
	In contrast, Our work focuses solely on optimizing the visual modality.

	We revisit 3D convolutions with the ViNet architecture~\cite{jain2021vinet},
	proposing ViNet-S, a computationally efficient model with a lightweight decoder
	using filter groups~\cite{8100116} and channel shuffle layers~\cite{zhang2018shufflenet},
	achieving a threefold reduction in size and parameters while improving SP
	performance. We also identify limitations in using action classification backbones
	like S3D~\cite{s3d}, which may miss background actions due to a focus on
	primary motion. Instead, we propose ViNet-A, leveraging Spatio-Temporal Action
	Localization (STAL)~\cite{acarnet, slowfast} with our lightweight decoder,
	which localizes and classifies actions within the scene, better capturing scene
	essence. ViNet-A excels, particularly in human-centric datasets like MVVA~\cite{mvva},
	by focusing on the most relevant features, such as the salient face in group settings.

	We further introduce ViNet-E, an ensemble of ViNet-S and ViNet-A, combining their
	strengths by averaging their predicted saliency maps. Despite its compact
	design, ViNet-E outperforms transformer-based approaches on various datasets without
	using audio cues. Our contributions include: 1) ViNet-S: A lightweight model with
	9 million parameters, surpassing the original ViNet~\cite{jain2021vinet} in
	performance. 2) ViNet-A: Utilizing a STAL backbone for enhanced performance in
	videos with multiple subjects. 3) ViNet-E: An ensemble of ViNet-S and ViNet-A,
	achieving SOTA results across multiple datasets. 4) Extensive experiments on nine
	datasets, providing qualitative and quantitative insights.
	% \begin{itemize}
	%     \item ViNet-S: A lightweight model with 9 million parameters, surpassing the original ViNet~\cite{jain2021vinet} in performance.
	%     \item ViNet-A: Utilizing a STAL backbone for enhanced performance in videos with multiple subjects.
	%     \item ViNet-E: An ensemble of ViNet-S and ViNet-A, achieving SOTA results across multiple datasets.
	%     \item Extensive experiments on nine datasets, providing qualitative and quantitative insights.
	% \end{itemize}

	%1) We propose ViNet-S, a lightweight and efficient model utilizing merely 9 million parameters in total while still achieving superior performance compared to ViNet~\cite{jain2021vinet}. 2) We also introduce ViNet-A, which utilizes a STAL backbone to enhance feature representation in videos with multiple subjects, resulting in significant performance improvements, especially in human-centric multi-person videos. 3) We empirically demonstrate that an ensemble of ViNet-S and ViNet-A can achieve significant gains over the individual models and obtain SOTA results on almost all studied datasets. 4) We present comprehensive qualitative and quantitative experiments utilizing nine different datasets.
	% \section{Proposed Model Architecture}

	\begin{figure*}[t!]
		\centering
		\includegraphics[width=\textwidth]{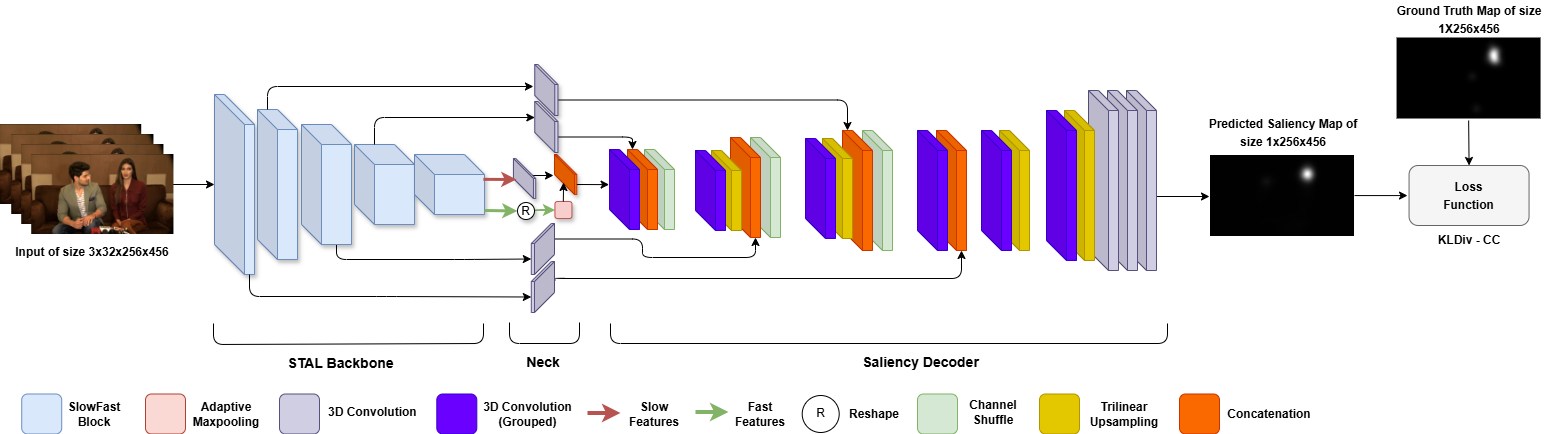}
		\caption{Our Model (ViNet-A) Architecture for SP (Best viewed in colour)}
		\label{fig:arch}
	\end{figure*}

	\section{Proposed Model Architecture}
	\label{sec:model_arch} We propose an end-to-end trainable visual-only model
	called ViNet-A (Figure \ref{fig:arch}). It is a fully 3D-convolutional encoder-decoder
	architecture consisting of a SlowFast network~\cite{slowfast} as the video encoder,
	a convolutional neck, and an efficient, lightweight decoder to reduce
	computational costs for predicting the saliency map. We also propose a variation
	of the ViNet architecture~\cite{jain2021vinet}, ViNet-S, which utilizes our efficient
	decoder, resulting in a small model while surpassing the original ViNet's
	performance. Lastly, we propose an ensemble of the two proposed models, ViNet-E.
	We elaborate on the proposed models in the following sections.
	\subsection{ViNet-A}
	\subsubsection{Backbone}

	Our model utilizes the SlowFast network~\cite{slowfast}, pre-trained on the AVA
	actions dataset~\cite{ava} as its video encoder. This backbone effectively
	captures localized actions across spatial and temporal dimensions. The SlowFast
	network comprises of two parallel pathways: the Slow pathway, which captures spatial
	semantics at a low frame rate, and the Fast pathway, which focuses on fine-grained
	temporal motion at a high frame rate.
	% Both pathways are 3D convolutional networks with lateral connections between them to combine information. These connections are then used as skip connections to the saliency decoder.
	Both pathways are 3D convolutional networks that combine information through
	lateral connections, which are subsequently used as skip connections in the saliency
	decoder.
	% The SlowFast network consists of two parallel pathways: the Slow and Fast pathways. The Slow pathway operates at a low frame rate and captures spatial semantics, while the Fast pathway operates at a high frame rate and captures motion at fine temporal resolution. Both pathways are 3D convolutional networks with varying input sizes. Each pathway produces features at different stages of the network. To combine information from both pathways, lateral connections are made from the Fast pathway to the Slow pathway at these stages. These lateral connections are then passed to different layers of the saliency decoder as skip connections.

	%, specifically pool1, res2, res3, res4, and res5

	% We input a video clip $\textit{x}_{clip} \in \mathbb{R}^{3 \times 32 \times 256 \times 456}$ to the encoder which outputs slow features, $X_{slow} \in \mathbb{R}^{2048 \times 8 \times 16 \times 29}$, fast features, $X_{fast} \in \mathbb{R}^{256 \times 32 \times 16 \times 29}$ and hierarchical features $X_{1}$, $X_{2}$, $X_{3}$ and $X_{4}$.
	% $X_{slow} \in \mathbb{R}^{2048 \times 8 \times 16 \times 29}$, fast features, $X_{fast} \in \mathbb{R}^{256 \times 32 \times 16 \times 29}$, and multi-scale features:

	% $X_{1} \in \mathbb{R}^{80 \times 8 \times 64 \times 114}$, $X_{2} \in \mathbb{R}^{320 \times 8 \times 64 \times 114}$

	% $X_{3} \in \mathbb{R}^{640 \times 8 \times 32 \times 57}$, $X_{4} \in \mathbb{R}^{1280 \times 8 \times 16 \times 29}$

	%-------------------------------------------------------------------------
	% \subsection{Neck}
	\subsubsection{Neck}

	% consequently decreasing the -> lowering
	The neck uses $1 \times 1$ convolutional blocks to reduce the number of
	channels, lowering computational overhead. We reduce the number of channels in
	$X_{slow}$ by half. $X_{fast}$ is reshaped to double its channels while halving
	its temporal dimension and then passed through an adaptive max pool to align its
	temporal dimension with $X_{slow}$. The two are concatenated channel-wise, resulting
	in fused SlowFast features,
	$X_{slowfast}\in \mathbb{R}^{1536 \times 8 \times 16 \times 29}$.
	% Hierarchical features $X_{1}$, $X_{2}$, $X_{3}$ and $X_{4}$ are also processed with $1 \times 1$ convolutional blocks to reduce their channels by half and improve computational efficiency. %In short, the process can be described as follows:
	Similarly, hierarchical features $X_{1}$, $X_{2}$, $X_{3}$, and $X_{4}$ are processed
	through $1 \times 1$ convolutional blocks to halve their channels, improving
	computational efficiency.

	% The neck fuses the slow and fast features and processes each of the Hierarchical features $X_{1}$, $X_{2}$, $X_{3}$ and $X_{4}$ before passing them to the saliency decoder. The slow and fast features are processed according to Equations \ref{eq:slow,eq:fast} for computational efficiency and concatenated to obtain
	% SlowFast features, $X_{slowfast} \in \mathbb{R}^{1536 \times 8 \times 16 \times 29}$. Hierarchial features are processed with
	% \begin{equation}
	%   X^{'}_{slow} = ReLU(Conv^{1 \times 1}(X_{slow}))
	%   \label{eq:slow}
	% \end{equation}
	% \begin{equation}
	%     X^{'}_{fast} = AdaptiveMaxPool(Reshape(X_{fast}))
	%     \label{eq:fast}
	% \end{equation}
	% \begin{equation}
	%     X_{slowfast} = [X^{'}_{slow}, X^{'}_{fast}]
	%     \label{eq:slowfast}
	% \end{equation}
	% \begin{equation}
	%     X_{i} = ReLU(Conv^{1 \times 1}(X_{i})), i \in {1, 2, 3, 4}
	%     \label{eq:Xi}
	% \end{equation}
	% Here, ReLU represents the ReLU activation function $Conv^{1 \times 1}$ represents ${1 \times 1}$ convolutions for halving the channels and $[,]$ represents concatenation.

	%-------------------------------------------------------------------------
	% \subsection{Saliency Decoder}
	\subsubsection{Saliency Decoder}
	% The Saliency Decoder comprises six decoding blocks consisting of 3D convolutions , trilinear upsampling and channel shuffle ~\cite{zhang2018shufflenet} layers. We use 3D convolutions with filter groups ~\cite{8100116} and channel shuffle to greatly reduce computation costs while maintaining accuracy. SlowFast features, $X_{slowfast}$, are input to the decoder and hierarchical features $X_{1}$, $X_{2}$, $X_{3}$, $X_{4}$ are passed as skip connections sequentially. All 3D convolutions except the last block utilize filter groups with 32, 16, 8, 8, 4 and 2 groups, respectively. Furthermore, channel shuffle layers are inserted after the first three grouped convolution layers. We experimented with different filter groups and channel shuffle layer configurations and found this optimal. We use the ReLU activation function after every convolutional layer except the last one, which employs the Sigmoid activation function to output the predicted saliency map.

	The Saliency Decoder consists of six decoding blocks with 3D convolutions
	using filter groups~\cite{8100116}, trilinear upsampling and channel shuffle~\cite{zhang2018shufflenet}
	layers to reduce computational costs while preserving accuracy. SlowFast
	features, $X_{slowfast}$, are fed into the decoder, with hierarchical features
	$X_{i}$ passed as skip connections. %. These design choices greatly reduce computation costs while maintaining accuracy. SlowFast features, $X_{slowfast}$, are input to the decoder and hierarchical features $X_{i}$ are passed as skip connections sequentially.
	All 3D convolutions, except the last block, utilize filter groups with 32, 16,
	8, 8, 4 and 2 groups, respectively, with channel shuffle layers applied after the
	first three grouped convolutions. %. Furthermore, channel shuffle layers are inserted after the first three grouped convolution layers.
	We experimented with different filter groups and channel shuffle layer configurations
	and found this setup optimal. ReLU activations follow every convolutional
	layer, except for the last, which uses Sigmoid to predict the saliency map. % We use the ReLU activation function after every convolutional layer except the last one, which employs the Sigmoid activation function to output the predicted saliency map.

	%-------------------------------------------------------------------------
	\subsection{ViNet-S \& ViNet-E}

	ViNet-S employs the S3D~\cite{s3d} backbone as its video encoder and the
	lightweight decoder with grouped convolutions and channel shuffle layers, similar
	to the ViNet-A saliency decoder described above.

	ViNet-E is an ensemble of the proposed models, ViNet-S and ViNet-A, which generates
	a saliency map by performing a simple pixel-wise mean of the two predicted saliency
	maps. Since both models predict saliency maps of different sizes, the ViNet-S prediction
	is upsampled to match ViNet-A before averaging. %we upsample the ViNet-S prediction to match that of ViNet-A before computing the pixel-wise mean.

	% \begin{equation}
	%     Sal_{E} = Mean(Sal_{A}, Upsample(Sal_{S}))
	% \end{equation}

	% Here, $Sal_{S}$, $Sal_{A}$ and $Sal_{E}$ represent the output saliency maps of ViNet-S, ViNet-A and ViNet-E, respectively.

	\section{Experiments}

	% Bhav
	% \subsection{Datasets}

	% We conducted experiments on three visual-only saliency datasets—DHF1K, Hollywood-2, and UCF-Sports—and six audio-visual saliency prediction datasets—AVAD, Coutrot1, Coutrot2, DIEM, ETMD, and MVVA. The \textbf{\textit{DHF1K}} dataset includes 1,000 videos, split into training, validation, and testing sets, though the test set's ground truth is unavailable. \textbf{\textit{Hollywood-2}}, the largest video SP dataset with 1,707 videos from Hollywood movies, was used with a standard training-test split. \textbf{\textit{UCF-Sports}} comprises 150 sports-related videos with a standard train-test split. \textbf{\textit{AVAD}} features 45 brief video clips of various scenes; \textbf{\textit{Coutrot1}} has 60 clips of natural scenes; \textbf{\textit{Coutrot2}} includes 15 clips of meetings with eye-tracking data; \textbf{\textit{DIEM}} contains 84 video clips from various genres; \textbf{\textit{ETMD}} consists of 12 videos from Hollywood movies; and \textbf{\textit{MVVA}}, a large-scale eye-tracking database, includes 300 multi-face videos across six categories, capturing eye movements of 34 subjects in diverse settings. For the experiments on audio-visual datasets we have used only the visual components.

	\paragraph{Datasets}
	We conduct experiments on three visual-only saliency datasets - DHF1K~\cite{dhf1k},
	Hollywood-2~\cite{hollywood-ucf}, and UCF-Sports~\cite{hollywood-ucf} and six
	audio-visual saliency prediction datasets- AVAD~\cite{avad}, Coutrot1~\cite{coutrot1conf,
	coutrot1journ}, Coutrot2~\cite{coutrot2}, DIEM~\cite{diem}, ETMD~\cite{etmd} and
	MVVA~\cite{mvva}.

	\begin{table*}
		[!t]
		\centering
		\caption{Quantitative comparison of model sizes and performance on visual-only
		datasets. }
		\begin{subtable}
			[t]{0.63\textwidth}
			\centering
			\caption{Results on DHF1K validation set, UCF-Sports and Hollywood2 test sets.
			Best results highlighted in red and second best in blue.}
			\label{table:1a}
			\resizebox{1.17\columnwidth}{!}{ %
			\begin{tabular}{@{}|l|llll|llll|llll|@{}}
				\hline
				\multirow{2}{*}{METHOD}         & \multicolumn{4}{|c|}{DHF1K (Validation Set)} & \multicolumn{4}{|c|}{UCF-Sports}   & \multicolumn{4}{|c|}{Hollywood2}    \\
				\cline{2-13}                    & CC$\uparrow$                                 & NSS$\uparrow$                      & AUC-J$\uparrow$                    & SIM$\uparrow$                      & CC$\uparrow$                       & NSS$\uparrow$                      & AUC-J$\uparrow$                    & SIM$\uparrow$                      & CC$\uparrow$                       & NSS$\uparrow$                      & AUC-J$\uparrow$                    & SIM$\uparrow$                      \\  \hline
				ACLNet~\cite{dhf1k}             & 0.434                                        & 2.35                               & 0.890                              & 0.315                              & 0.510                              & 2.56                               & 0.897                              & 0.406                              & 0.623                              & 3.08                               & 0.913                              & 0.542                              \\  \hline
				TASED-Net~\cite{min2019tased}   & 0.481                                        & 2.706                              & 0.894                              & 0.362                              & 0.582                              & 2.920                              & 0.899                              & 0.469                              & 0.646                              & 3.302                              & 0.918                              & 0.507                              \\  \hline
				UNISAL~\cite{droste2020unified} & 0.490                                        & 2.77                               & 0.901                              & 0.390                              & 0.644                              & 3.38                               & 0.918                              & 0.523                              & 0.673                              & 3.90                               & 0.934                              & 0.542                              \\  \hline
				ViNet~\cite{jain2021vinet}      & 0.521                                        & 2.957                              & 0.919                              & 0.388                              & 0.673                              & 3.620                              & 0.924                              & 0.522                              & 0.693                              & 3.730                              & 0.930                              & 0.550                              \\  \hline
				TSFP-Net~\cite{tsfpnet}         & 0.529                                        & 3.009                              & 0.919                              & 0.397                              & 0.685                              & 3.698                              & 0.923                              & 0.561                              & 0.711                              & 3.910                              & 0.936                              & 0.571                              \\  \hline
				STSANet~\cite{stsanet}          & 0.539                                        & 3.082                              & 0.920                              & 0.411                              & 0.721                              & 3.927                              & 0.936                              & 0.560                              & 0.705                              & 3.908                              & 0.938                              & 0.579                              \\  \hline
				TMFI-Net~\cite{tmfinet}         & \textcolor{red}{\underline{0.554}}           & \textcolor{red}{\underline{3.201}} & \textcolor{red}{\underline{0.924}} & \textcolor{red}{\underline{0.428}} & 0.707                              & 3.863                              & 0.936                              & 0.565                              & 0.739                              & 4.095                              & 0.940                              & 0.607                              \\  \hline
				THTD-Net~\cite{thtdnet}         & \textcolor{blue}{0.553}                      & \textcolor{blue}{3.188}            & \textcolor{red}{\underline{0.924}} & \textcolor{blue}{0.425}            & 0.711                              & 3.840                              & 0.933                              & 0.565                              & 0.726                              & 3.965                              & 0.939                              & 0.585                              \\  \hline
				DiffSal~\cite{diffsal}          & 0.533                                        & 3.066                              & 0.918                              & 0.405                              & 0.685                              & 3.483                              & 0.928                              & 0.543                              & \textcolor{blue}{0.765}            & 3.955                              & \textcolor{red}{\underline{0.951}} & \textcolor{red}{\underline{0.610}} \\  \hline
				\hline
				ViNet-S                         & 0.529                                        & 3.008                              & 0.919                              & 0.399                              & 0.673                              & 3.652                              & 0.930                              & 0.530                              & 0.728                              & 3.941                              & 0.941                              & 0.582                              \\  \hline
				ViNet-A                         & 0.525                                        & 3.019                              & 0.916                              & 0.399                              & \textcolor{blue}{0.734}            & \textcolor{blue}{4.108}            & \textcolor{blue}{0.940}            & \textcolor{blue}{0.586}            & 0.756                              & \textcolor{blue}{4.119}            & 0.945                              & 0.604                              \\  \hline
				ViNet-E                         & 0.549                                        & 3.134                              & \textcolor{blue}{0.922}            & 0.409                              & \textcolor{red}{\underline{0.744}} & \textcolor{red}{\underline{4.156}} & \textcolor{red}{\underline{0.941}} & \textcolor{red}{\underline{0.587}} & \textcolor{red}{\underline{0.766}} & \textcolor{red}{\underline{4.168}} & \textcolor{blue}{0.947}            & \textcolor{blue}{0.609}            \\  \hline
			\end{tabular}
			}
		\end{subtable}
		\hfill
		\begin{subtable}
			[t]{0.258\textwidth}
			\begin{flushright}
				\caption{Quantitative comparison of model sizes \& parameters}
				\resizebox{\columnwidth}{!}{ %
				\begin{tabular}{@{}|l|P{1.5cm}|P{1.5cm}|@{}} %P{0.85cm}
					\hline
					Model                           & Size (MB) & \# Params (Million) \\  \hline
					ACLNet~\cite{dhf1k}             & 250       & 65.54               \\  \hline
					TASED-Net~\cite{min2019tased}   & 82        & 21.5                \\  \hline
					STAViS~\cite{tsiami2020stavis}  & 79.19     & 20.76               \\  \hline
					UNISAL~\cite{droste2020unified} & 15.5      & 4.06                \\  \hline
					ViNet~\cite{jain2021vinet}      & 124       & 32.5                \\  \hline
					TSFP-Net~\cite{tsfpnet}         & 58.4      & 15.3                \\ \hline
					STSA-Net~\cite{stsanet}         & 643       & 168.56              \\  \hline
					TMFI-Net~\cite{tmfinet}         & 234       & 61.34               \\  \hline
					THTD-Net~\cite{thtdnet}         & 220       & 57.67               \\  \hline
					CASP-Net~\cite{xiong2023casp}   & 196.91    & 51.62               \\  \hline
					DiffSal~\cite{diffsal}          & 269       & 70.54               \\  \hline
					\hline
					ViNet-S                         & 36.24     & 9.5                 \\  \hline
					ViNet-A                         & 147.6     & 38.69               \\  \hline
					ViNet-E                         & 183.84    & 48.19               \\  \hline
				\end{tabular}

				\label{table:sizes} }
			\end{flushright}
		\end{subtable}

		\label{table:1}
	\end{table*}

	\begin{table*}
		[!t]
		\centering
		\caption{Quantitative comparison results on the AVAD, Coutrot1, Coutrot2 and
		ETMD test sets.}
		\resizebox{\textwidth}{!}{ %
		\begin{tabular}{@{}|l|llll|llll|llll|llll|@{}}
			\hline
			\multirow{2}{*}{METHOD}        & \multicolumn{4}{|c|}{Coutrot1}     & \multicolumn{4}{|c|}{Coutrot2}     & \multicolumn{4}{|c|}{ETMD}         & \multicolumn{4}{|c|}{AVAD}          \\ %& \multicolumn{4}{|c|}{ADDED-COL} \\ % Add new column header here
			\cline{2-17}                    % Adjust the range to cover the new columns
			                               & CC$\uparrow$                       & NSS$\uparrow$                      & AUC-J$\uparrow$                    & SIM$\uparrow$                      & CC$\uparrow$                       & NSS$\uparrow$                    & AUC-J$\uparrow$                    & SIM$\uparrow$                      & CC$\uparrow$                       & NSS$\uparrow$                      & AUC-J$\uparrow$                    & SIM$\uparrow$                      & CC$\uparrow$                       & NSS$\uparrow$                      & AUC-J$\uparrow$                    & SIM$\uparrow$                      \\
			% & CC$\uparrow$                      & NSS$\uparrow$                    & AUC-J$\uparrow$                   & SIM$\uparrow$   \\ % Add metrics for the new column
			\hline
			ACLNet~\cite{dhf1k}            & 0.425                              & 1.92                               & 0.85                               & 0.361                              & 0.448                              & 3.16                             & 0.926                              & 0.322                              & 0.477                              & 2.36                               & 0.915                              & 0.329                              & 0.580                              & 3.17                               & 0.905                              & 0.446                              \\ %& 0.400 & 1.90 & 0.84 & 0.355 \\ % Add new row data here
			\hline
			TASED-Net~\cite{min2019tased}  & 0.479                              & 2.18                               & 0.867                              & 0.388                              & 0.437                              & 3.17                             & 0.921                              & 0.314                              & 0.509                              & 2.63                               & 0.916                              & 0.366                              & 0.601                              & 3.16                               & 0.914                              & 0.439                              \\ %& 0.469 & 2.20 & 0.870 & 0.389 \\ % Add new row data here
			\hline
			STAViS~\cite{tsiami2020stavis} & 0.458                              & 1.99                               & 0.861                              & 0.384                              & 0.652                              & 4.19                             & 0.940                              & 0.447                              & 0.560                              & 2.84                               & 0.929                              & 0.412                              & 0.604                              & 3.07                               & 0.915                              & 0.443                              \\ %& 0.654 & 4.20 & 0.945 & 0.450 \\ % Add new row data here
			\hline
			ViNet~\cite{jain2021vinet}     & 0.551                              & 2.68                               & 0.886                              & 0.423                              & 0.724                              & 5.61                             & 0.95                               & 0.466                              & 0.569                              & 3.06                               & 0.928                              & 0.409                              & 0.694                              & 3.82                               & 0.928                              & 0.504                              \\ %& 0.558 & 3.05 & 0.930 & 0.415 \\ % Add new row data here
			\hline
			TSFP-Net~\cite{tsfpnet}        & 0.57                               & 2.75                               & 0.894                              & 0.451                              & 0.718                              & 5.30                             & 0.957                              & 0.516                              & 0.576                              & 3.09                               & 0.932                              & 0.433                              & 0.688                              & 3.79                               & 0.932                              & 0.530                              \\ %& 0.575 & 3.10 & 0.935 & 0.430 \\ % Add new row data here
			\hline
			CASP-Net~\cite{xiong2023casp}  & 0.561                              & 2.65                               & 0.889                              & 0.456                              & 0.788                              & 6.34                             & 0.963                              & 0.585                              & 0.620                              & 3.34                               & 0.940                              & \textcolor{red}{\underline{0.478}} & 0.691                              & 3.81                               & 0.933                              & 0.528                              \\ %& 0.630 & 3.35 & 0.945 & 0.480 \\ % Add new row data here
			\hline
			\hline
			ViNet-S                        & 0.574                              & 2.876                              & 0.898                              & 0.449                              & 0.754                              & 6.103                            & 0.958                              & 0.547                              & 0.599                              & 3.268                              & \textcolor{blue}{0.941}            & 0.458                              & \textcolor{blue}{0.712}            & 4.090                              & \textcolor{blue}{0.935}            & \textcolor{blue}{0.540}            \\ %& 0.605                        & 3.275                        & 0.942                        & 0.459                        \\ % Add new row data here
			\hline
			ViNet-A                        & \textcolor{blue}{0.600}            & \textcolor{blue}{3.033}            & \textcolor{blue}{0.900}            & \textcolor{blue}{0.459}            & \textcolor{red}{\underline{0.862}} & \textcolor{red}{\underline{6.8}} & \textcolor{blue}{0.961}            & \textcolor{red}{\underline{0.638}} & \textcolor{blue}{0.623}            & \textcolor{blue}{3.379}            & \textcolor{blue}{0.941}            & 0.458                              & 0.709                              & \textcolor{blue}{4.094}            & 0.933                              & 0.534                              \\ %& \textcolor{blue}{0.625}  & \textcolor{blue}{3.380}  & \textcolor{blue}{0.943}   & \textcolor{blue}{0.460} \\ % Add new row data here
			\hline
			ViNet-E                        & \textcolor{red}{\underline{0.614}} & \textcolor{red}{\underline{3.085}} & \textcolor{red}{\underline{0.905}} & \textcolor{red}{\underline{0.465}} & \textcolor{blue}{0.854}            & \textcolor{blue}{6.762}          & \textcolor{red}{\underline{0.962}} & \textcolor{blue}{0.628}            & \textcolor{red}{\underline{0.632}} & \textcolor{red}{\underline{3.437}} & \textcolor{red}{\underline{0.943}} & \textcolor{blue}{0.468}            & \textcolor{red}{\underline{0.729}} & \textcolor{red}{\underline{4.167}} & \textcolor{red}{\underline{0.938}} & \textcolor{red}{\underline{0.547}} \\ %& \textcolor{red}{\underline{0.635}}  & \textcolor{red}{\underline{3.450}}  & \textcolor{red}{\underline{0.945}}   & \textcolor{red}{\underline{0.470}} \\ % Add new row data here
			\hline
		\end{tabular}
		}
		\label{table:2}
	\end{table*}

	% \vspace{10pt}

	\begin{table*}
		[!t]
		\centering
		\caption{Quantitative comparison results on the DIEM and MVVA test sets.}
		\begin{subtable}
			[t]{0.49\textwidth}
			\centering
			% \caption{Results on DIEM dataset.}
			\resizebox{0.72\columnwidth}{!}{ %
			\begin{tabular}{@{}|l|llll|@{}}
				\hline
				\multirow{2}{*}{METHOD}        & \multicolumn{4}{|c|}{DIEM}          \\
				\cline{2-5}                    & CC$\uparrow$                       & NSS$\uparrow$                      & AUC-J$\uparrow$                    & SIM$\uparrow$                      \\
				\hline
				ACLNet~\cite{dhf1k}            & 0.522                              & 2.02                               & 0.869                              & 0.427                              \\
				\hline
				TASED-Net~\cite{min2019tased}  & 0.557                              & 2.16                               & 0.881                              & 0.461                              \\
				\hline
				STAViS~\cite{tsiami2020stavis} & 0.579                              & 2.26                               & 0.883                              & 0.482                              \\
				\hline
				ViNet~\cite{jain2021vinet}     & 0.626                              & 2.47                               & 0.898                              & 0.483                              \\
				\hline
				TSFP-Net~\cite{tsfpnet}        & 0.651                              & 2.62                               & 0.906                              & 0.527                              \\
				\hline
				CASP-Net~\cite{xiong2023casp}  & 0.655                              & 2.61                               & 0.906                              & 0.543                              \\
				\hline
				\hline
				ViNet-S                        & 0.673                              & 2.732                              & \textcolor{blue}{0.908}            & 0.533                              \\
				\hline
				ViNet-A                        & \textcolor{blue}{0.675}            & \textcolor{blue}{2.742}            & \textcolor{blue}{0.908}            & \textcolor{blue}{0.547}            \\
				\hline
				ViNet-E                        & \textcolor{red}{\underline{0.701}} & \textcolor{red}{\underline{2.840}} & \textcolor{red}{\underline{0.913}} & \textcolor{red}{\underline{0.566}} \\
				\hline
			\end{tabular}
			}
		\end{subtable}
		\hfill
		\begin{subtable}
			[t]{0.49\textwidth}
			\centering
			% \caption{Results on MVVA dataset.}
			\resizebox{0.8\columnwidth}{!}{ %
			\begin{tabular}{@{}|l|llll|@{}}
				\hline
				\multirow{2}{*}{METHOD}        & \multicolumn{4}{|c|}{MVVA}          \\
				\cline{2-5}                    & CC$\uparrow$                       & NSS$\uparrow$                      & AUC-J$\uparrow$                    & KLDIV$\downarrow$                  \\
				\hline
				VASM~\cite{mvva}               & 0.722                              & 3.976                              & 0.905                              & 0.823                              \\
				\hline
				VAM-Net~\cite{vamnet}          & 0.741                              & 4.002                              & 0.912                              & 0.783                              \\
				\hline
				TASED-Net~\cite{min2019tased}  & 0.653                              & 3.319                              & 0.905                              & 0.970                              \\
				\hline
				STAViS~\cite{tsiami2020stavis} & 0.77                               & 3.060                              & 0.91                               & 0.80                               \\
				\hline
				ViNet~\cite{jain2021vinet}     & 0.81                               & 4.470                              & 0.93                               & 0.75                               \\
				\hline
				\hline
				ViNet-S                        & 0.802                              & 4.617                              & 0.933                              & 0.715                              \\
				\hline
				ViNet-A                        & \textcolor{blue}{0.825}            & \textcolor{red}{\underline{4.823}} & \textcolor{blue}{0.934}            & \textcolor{blue}{0.678}            \\
				\hline
				ViNet-E                        & \textcolor{red}{\underline{0.828}} & \textcolor{blue}{4.816}            & \textcolor{red}{\underline{0.936}} & \textcolor{red}{\underline{0.663}} \\
				\hline
			\end{tabular}
			}
		\end{subtable}

		\label{table:3}
	\end{table*}

	\paragraph{Training}
	Following~\cite{jain2021vinet}, we input a clip of 32 consecutive frames to the
	ViNet-S model and use the ground truth saliency map of the 32\textsuperscript{nd}
	frame for supervision and prediction. For the ViNet-A model, the input consists
	of 32 frames sampled from a window of 64 consecutive frames by selecting every
	alternate frame. We use the ground truth saliency map of the 33\textsuperscript{rd}
	frame for supervision and prediction, akin to action label predictions in STAL
	models \cite{acarnet}. Both models are trained using the Adam optimizer with a
	learning rate of $10^{-4}$ and batch size of 8 for ViNet-S and 6 for ViNet-A.

	For evaluating our model on DHF1K, we use the validation set due to unavailable
	annotations for the test set, as in prior efforts~\cite{tinyhd, ma2022video}. We
	use the standard train and test sets provided for training on datasets Hollywood-2,
	UCF-Sports and DIEM. For Coutrot1, Coutrot2, AVAD, and ETMD, we perform 3-fold
	cross-validation and report average metrics across the splits. For MVVA, we follow~\cite{vamnet}
	and perform training on a random split.

	%We use the validation/test set of the datasets for early stopping.

	% \subsection{Evaluation Metrics}

	% We evaluate our method on five standard evaluation metrics \cite{metrics}. Specifically, we employ the following metrics: \textit{Correlation Coefficient} (CC) calculates the Pearson correlation between the ground truth and the predicted saliency maps. \textit{Similarity} (SIM) measures the overlap between the predicted and ground truth saliency maps by computing the histogram intersection. \textit{Area Under the Receiver Operating Characteristic Curve} (AUC-J), treats the saliency map as a binary classifier and evaluates its performance across different threshold values. A ROC curve is generated by plotting the true positive rate against the false positive rate at varying thresholds. \textit{Normalized Scanpath Saliency} (NSS) assesses the average normalized saliency at fixated locations. \textit{Kullback-Leibler Divergence} (KLDiv) Loss measures the difference between the predicted and ground truth saliency distributions using an information-theoretic approach.

	\paragraph{Evaluation Metrics}
	We evaluate our method on five standard evaluation metrics whose details can be
	found in \cite{metrics}: AUC-Judd (AUC-J), Similarity Metric (SIM),
	Correlation Coefficient(CC), Normalized Scanpath Saliency(NSS) and Kullback-Leibler
	Divergence(KLDiv). Except for KLDiv, higher metric values indicate better
	model performance.%Except for KLDiv, the larger the metric value, the better the model's performance.

	%------------------------------------------------------------------------------------------
	% \subsection{Loss Function}
	\paragraph{Loss Function}
	We utilize a combination of the above evaluation metrics, a standard technique
	in saliency tasks \cite{metrics}.
	% After experimenting with various combinations, we found that for most datasets, the best results were achieved using the loss function:  \(Loss = KLDiv(P,Q) - CC(P,Q)\), where $P$ \& $Q$ are the predicted saliency map and ground truth respectively.
	Through experimentation with different combinations, we found that the optimal
	results for most datasets were achieved with the loss function:
	$Loss = KLDiv(P, Q) - CC(P, Q)$, where $P$ \& $Q$ represent the predicted saliency
	map and ground truth, respectively.

	% \begin{equation}
	%   Loss = KLDiv(P, Q) - CC(P, Q)
	%   \label{eq:loss_function}
	% \end{equation}

	% % \[Loss = KLDiv(P, Q) - CC(P, Q)\]
	% \begin{equation}
	%   KLDiv(P,Q)=\sum\limits_i P_i\log(\epsilon+\frac{Q_i}{P_i+\epsilon})
	%   \label{eq:kldiv_loss}
	% \end{equation}

	% \begin{equation}
	%   CC(P,Q)=\frac{\sigma(P,Q)}{\sigma(P,P) \times \sigma(Q,Q)}
	%   \label{eq:cc_loss}
	% \end{equation}

	% where $P$ \& $Q$ are the predicted saliency map and ground truth respectively, $\epsilon$ is a regularization term and \( \sigma(P,Q)\) represents covariance between \(P\) and \(Q\)
	%----------------------------------------------------------------------------

	%-------------------------------------------------------
	\section{Results and Discussions}
	We evaluate the proposed models by comparing them against thirteen different
	methods from previous research. These include four 3D convolution-based approaches:
	ViNet~\cite{jain2021vinet}, TASED-Net~\cite{min2019tased}, STAVIS~\cite{tsiami2020stavis},
	and TSFP-Net~\cite{tsfpnet}; two methods utilizing recurrent networks: ACLNet~\cite{dhf1k}
	and UNISAL~\cite{droste2020unified}; four models employing transformers: STSANet~\cite{stsanet},
	THTD-Net~\cite{thtdnet}, CASP-Net~\cite{xiong2023casp}, TMFI-Net~\cite{tmfinet};
	one diffusion-based model: DiffSal~\cite{diffsal} and a couple of multi-branch
	network methods: VAM-Net~\cite{vamnet} and VASM~\cite{mvva}. Six of these
	models (STAVIS, CASP-Net, TSFP-Net, VAM-Net, DiffSal and VASM) additionally
	employ audio information in their approach. We report results directly from the
	corresponding papers when available. If the code is publicly available and
	executable, we compute their results on other datasets.

	\paragraph{Visual Only Datasets}
	Table \ref{table:1a} presents results on the visual-only datasets. The model
	sizes and the number of parameters of the studied models are presented in
	Table \ref{table:sizes}. We observe that ViNet-E achieves the best performance
	on UCF-Sports and Hollywood2 datasets~\cite{hollywood-ucf}, while achieving
	competent results on the DHF1K dataset~\cite{dhf1k}. Interestingly, ViNet-A
	also outperforms the previous methods on the UCF-Sports and Hollywood2
	datasets. Its strong performance on these two human-centric datasets clearly demonstrates
	the advantages of using an STAL backbone over an action classification backbone.
	Notably, all three proposed models, including ViNet-S, consistently surpass the
	base ViNet model. %Another notable observation is that all three proposed models, including ViNet-S, consistently outperform the base ViNet model.

	The ViNet-S model recovers most of the underlying performance in all the cases
	while using only a tiny fraction of the parameters. For instance, on the
	Hollywood2 dataset, the largest SP dataset with 884 videos in the test set,
	ViNet-S recovers over 98.5\% performance on the CC metric compared to the transformer-based
	SOTA TMFI-Net, while bringing over six-fold reduction in terms of number of
	parameters (Table \ref{table:1}). Interestingly, ViNet-S outperforms TMFI-Net
	on the AUC-J metric. UNISAL is the only model lighter than the ViNet-S model.
	However, it consistently underperforms in comparison, possibly due to its
	recurrent architecture.

	\paragraph{Audio Visual Datasets}
	Table \ref{table:2} and Table \ref{table:3} present results on the audio-visual
	datasets. The proposed ViNet-S, ViNet-A, and ViNet-E consistently outperform
	prior models across all six datasets, consistently ranking among the top two models.
	The videos in the Coutrot2 and MVVA datasets emphasize multi-person
	interactions. Notably, ViNet-A achieves significant improvements on both
	datasets, maintaining a consistent performance trend with the other human-centric
	datasets. On MVVA (the largest audio-visual saliency dataset), while only using
	the visual modality ViNet-A brings over 20\% gains on NSS metric over the
	complex multi-branch VAM-Net, which uses an explicit combination of motion,
	texture, face and audio features.

	On four out of the six audio-visual datasets, i.e. DIEM, AVAD, Coutrot1, and
	MVVA, the smaller ViNet-S surpasses all the previous methods. The consistent
	performance improvements of the ViNet-E model validate the effectiveness of
	the proposed ensemble strategy, establishing a new SOTA in most datasets.
	Another notable observation is that incorporating audio information does not
	appear to provide a significant advantage for the task of SP. Consistent with prior
	studies~\cite{agrawal2022does,jain2021vinet,tsfpnet}, we found that several
	audio-visual models~\cite{diffsal,tsiami2020stavis}, in reality, are not
	exploiting the audio information. At inference, the models appear agnostic to
	the audio information, i.e., the results remain the same irrespective of sending
	the random audio or zero audio. This represents a significant scientific flaw
	that requires further investigation in future research, and comparisons with their
	results should be approached with caution. Although ViNet-E outperforms their
	audio-visual version on several datasets, we limit our comparisons only to their
	visual only model.

	% \begin{figure*}[t!]
	%   \centering
	%   \includegraphics[width=0.9\columnwidth]{Images/qual_v2.drawio.png}
	%   \caption{Qualitative results: Comparing Ground Truth Saliency Maps with the predicted output of our model and previous SOTA models ViNet~\cite{jain2021vinet} and STAViS~\cite{tsiami2020stavis} on four different datasets - AVAD~\cite{avad}, Coutrot1~\cite{coutrot1conf, coutrot1journ}, MVVA~\cite{mvva}, and Coutrot2~\cite{coutrot2}}
	%   \label{fig:qual}

	% \end{figure*}

	% \begin{figure}[t!]
	%   \centering
	%   \includegraphics[width=\columnwidth]{Images/qual_v2.drawio.png}
	%   \caption{Qualitative results: Comparing Ground Truth Saliency Maps with the predicted output of our model and previous SOTA model ViNet and STAViS on four different datasets - AVAD, Coutrot1, MVVA, and Coutrot2.}
	%   \label{fig:qual}
	% \end{figure}

	\begin{figure}[t!]
		\centering
		\includegraphics[width=\columnwidth]{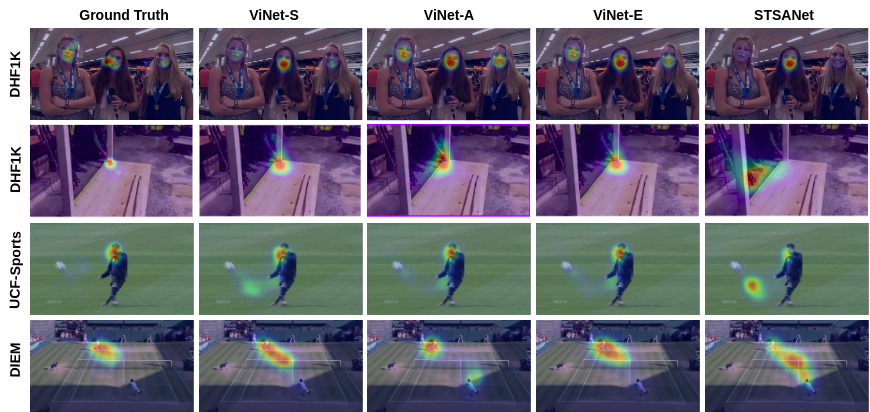}
		\caption{Qualitative results: Comparing Ground Truth with the predicted saliency
		maps of our models and STSANet on three different datasets - DHF1K, UCF-Sports
		and DIEM.}
		\label{fig:qual}
	\end{figure}

	\paragraph{Qualitative comparisons}
	Figure \ref{fig:qual} shows the qualitative performance of our ViNet-S, ViNet-A
	and ViNet-E models on video sequences from three different datasets: DHF1K, UCF-Sports
	and DIEM. We observe that STAL features efficiently capture the interaction
	between an actor/object with the context (surrounding) as evident in the
	strong performance of our model. ViNet-E is consistently closer to the ground truth
	in different settings than all other models, including STSANet. %ViNet-E incorporates both global and localized action features to better model the saliency and is consistently closer to the ground truth in different scenarios.
	%We observe that ViNet-E is consistently closer to the ground truth in different settings than all other models, including STSANet. % We obseve that ViNet-A is more closer to the ground truth in multi-person scenario

	% shows the qualitative performance of our ViNet-A model on video sequences from four different datasets: AVAD, Coutrot1, MVVA, and Coutrot2. We observe that the STAL features efficiently capture the interaction between an actor/object with the context (surroundings), as evident in the strong performance of our model. It also retains performance in videos where humans are absent. This observation underscores our model's capacity to comprehensively grasp the entirety of a scene, including the intricate interplay among various objects and participants. This stands in contrast to the limitations exhibited by single-label backbone models (as observed in samples from Coutrot2 and AVAD datasets).

	\paragraph{Computational load}
	Table \ref{table:sizes} compares the different models in terms of models size and
	number of parameters. The proposed decoder significantly reduces the number of
	parameters in ViNet-S compared to the original ViNet model. Aside from UNISAL,
	ViNet-S is the most efficient in terms of model size and parameters among the
	compared models. % We observe that the proposed decoder brings significant parameter reduction in ViNet-S, compared to the original ViNet model. Except for UNISAL, ViNet-S fares best among the other models in terms of both the model size and number of parameters.
	We observe that switching from the S3D backbone in ViNet-S to the SlowFast
	backbone in ViNet-A leads to significant parameter gains. Notably, ViNet-A's
	decoder contains only 1.6 million parameters, while the SlowFast backbone accounts
	for the remaining 37 million. Lastly, the ViNet-E model remains smaller than state-of-the-art
	transformer-based models (e.g., TMFI-Net and THTD-Net) in both model size and parameter
	count.
	% It is worth noting that the decoder of ViNet-A comprises only 1.6 million parameters, while the remaining 37 million parameters are contributed by the SlowFast backbone. Finally, the ViNet-E model remains below the SOTA transformer-based models (e.g. TFMI-Net, and THTD-Net), both in terms of model size and parameters.

	The non-autoregressive design of the proposed ViNet models enables parallel processing,
	providing a significant advantage over autoregressive models such as UNISAL, which
	rely on frame-level recurrence. On an Nvidia RTX 4090 GPU, ViNet-S, ViNet-A, and
	ViNet-E models achieve runtimes of approximately 200fps, 120fps, and 90fps,
	respectively, in a real-time processing setup (with a batch size of one). With
	a batch size of eight, ViNet-S reaches an impressive 1070fps. %the ViNet-S model achieves an impressive runtime performance of 1070fps.

	\section{Conclusion}
	% This work presents two efficient models, ViNet-S and ViNet-A, based on simple architecture design choices. ViNet-S is lightweight yet retains or improves upon most of the convolutional methods. Considering the complementary nature of information inherent in action classification and action detection, we propose ViNet-A, achieving SOTA results on visual as well as audio-visual datasets even without utilizing audio cues. Interestingly, an ensemble of straightforward pixel-wise mean of predictions from the two proposed models further pulls away from the competition. This opens up a new avenue for integrating the global and localized features from action classification and detection. Furthermore, in the current work we only tackle model optimization from the perspective of model architecture, future works should also address this from the lens of model compression and knowledge distillation.

	This work introduces two efficient models, ViNet-S and ViNet-A, characterized by
	their simple architectural design choices. %built on simple architectural design choices.
	ViNet-S is lightweight yet matches or surpasses most convolutional methods,
	while ViNet-A, which utilizes localized action features, consistently performs
	well on human-centric datasets with multiple subjects. ViNet-E, the ensemble
	model, leverages the complementary nature of action classification and detection
	to achieve state-of-the-art results on both visual and audio-visual datasets,
	even without audio cues. Using pixel-wise averaging enhances performance,
	suggesting new avenues for integrating global and localized action features.
	% While this study focuses on model optimization through architectural enhancements, future efforts should explore model compression and knowledge distillation.
	While this study emphasizes model optimization primarily through architectural
	refinements, future work would aim to investigate and integrate ideas from
	model compression and knowledge distillation methodologies.

	% ViNet-E leverages the complementary nature of action classification and detection to achieve state-of-the-art results on both visual and audio-visual datasets, even without audio cues. An ensemble of the two models using pixel-wise averaging further enhances performance, suggesting new avenues for integrating global and localized action features. %opening new possibilities for integrating global and localized action features.

	%%%%%%%%% REFERENCES
	\bibliographystyle{IEEEtran}
	\bibliography{reference}
\end{document}